%% file: CVPR_2026.tex
\documentclass[10pt,twocolumn,letterpaper]{article}

\usepackage{cvpr}              
\usepackage[accsupp]{axessibility}  

\usepackage{url}            
\usepackage{booktabs}       
\usepackage{amsfonts}       
\usepackage{nicefrac}       
\usepackage{microtype}      
\usepackage{xcolor}         
\usepackage{graphicx}
\usepackage{algorithm}
\usepackage{algorithmic}

\usepackage{listings}

\usepackage{tikz}
\usetikzlibrary{positioning, arrows.meta, shapes}

\input{preamble}

\definecolor{cvprblue}{rgb}{0.21,0.49,0.74}
\usepackage[pagebackref,breaklinks,colorlinks,allcolors=cvprblue]{hyperref}

\def\paperID{33805} 
\def\confName{CVPR}
\def\confYear{2026}

\title{Consistency Beyond Contrast: Enhancing Open-Vocabulary Object Detection Robustness via Contextual Consistency Learning}

\author{
Bozhao Li$^{1,2}$, Shaocong Wu$^{2}$, Tong Shao$^{1}$, Senqiao Yang$^{3}$\\
Qiben Shan$^{2}$, Zhuotao Tian$^{1}$\thanks{Corresponding authors.}, Jingyong Su$^{1,2}$\footnotemark[1]\\
$^{1}$Harbin Institute of Technology, Shenzhen \quad
$^{2}$Pengcheng Laboratory \quad
$^{3}$CUHK\\
{\tt\small \{23B951035, shaotong\}@stu.hit.edu.cn, \{wushc, shanqb\}@pcl.ac.cn}\\
{\tt\small senqiaoyang@link.cuhk.edu.hk, \{tianzhuotao, sujingyong\}@hit.edu.cn}
}

\begin{document}
\maketitle
\input{sec/0_abstract}

\input{sec/1_introduction}
\input{sec/2_Relatedwork}

\input{sec/3_Method}

\input{sec/4_experiment}

\input{sec/5_conclusion}
\input{sec/ack}

{
    \small
    \bibliographystyle{ieeenat_fullname}
    \bibliography{main}
}


\input{sec/X_suppl}

\end{document}

%% file: preamble.tex
\usepackage{amsmath}
\usepackage{multirow}
\usepackage{caption}
\usepackage{listings}
\usepackage{enumitem}

%% file: sec/0_abstract.tex
\begin{abstract}
Recent advances in open-vocabulary object detection focus primarily on two aspects: scaling up datasets and leveraging contrastive learning to align language and vision modalities. However, these approaches often neglect internal consistency within a single modality, particularly when background or environmental changes occur.  This lack of consistency leads to a performance drop because the model struggles to detect the same object in different scenes, which reveals a robustness gap. To address this issue, we introduce Contextual Consistency Learning (CCL), a novel framework that integrates two key strategies: Contextual Bootstrapped Data Generation (CBDG) and Contextual Consistency Loss (CCLoss). CBDG functions as a data generation mechanism, producing images that contain the same objects across diverse backgrounds. This is essential because existing datasets alone do not support our CCL framework. The CCLoss further enforces the invariance of object features despite environmental changes, thereby improving the model's robustness in different scenes. These strategies collectively form a unified framework for ensuring contextual consistency within the same modality. Our method achieves state-of-the-art performance, surpassing previous approaches by +16.3 AP on OmniLabel and +14.9 AP on $D^3$. These results demonstrate the importance of enforcing intra-modal consistency, significantly enhancing model generalization in diverse environments. Our code is publicly available at: \href{https://github.com/bozhao-li/CCL}{https://github.com/bozhao-li/CCL}.
\end{abstract}

%% file: sec/1_introduction.tex
\section{Introduction}
\label{sec:intro}

\begin{figure*}[t] 
    \centering
    \includegraphics[width=\textwidth]{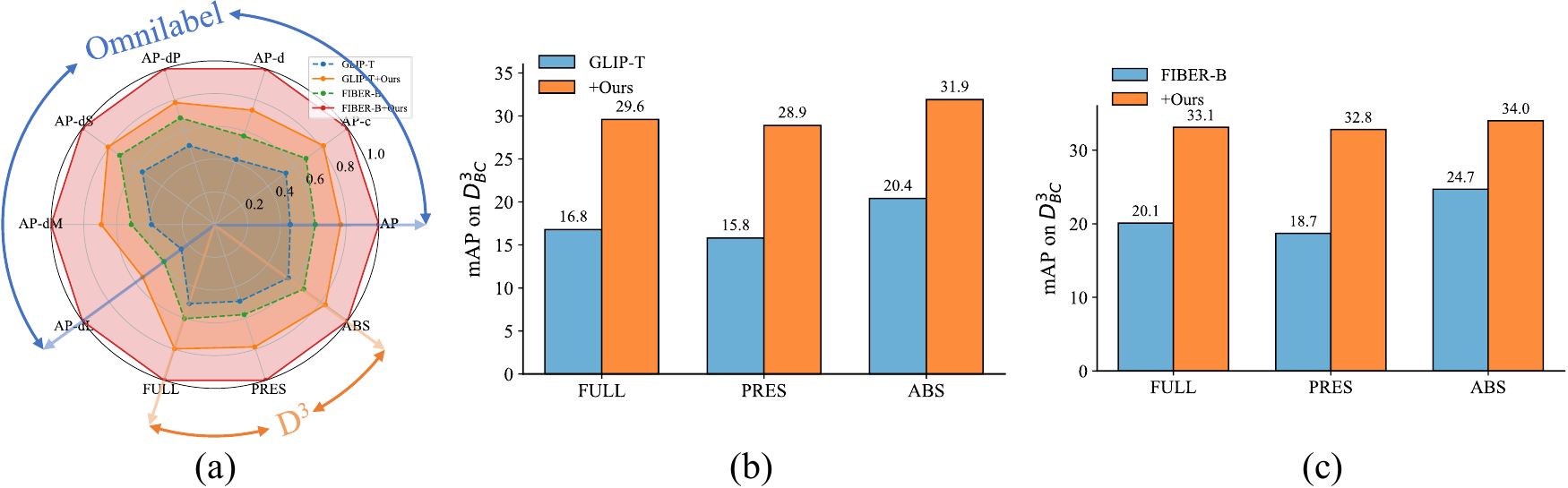} 
    \caption{Performance and robustness comparison of different methods. (a) Our approach, with Contextual Consistency Learning, achieves the best overall results, reaching a normalized score of 1 in all metrics. (b,c) Benchmark backgrounds are altered to test robustness. Tested on $D^3$\textsubscript{BC}, baseline methods degrade, while ours remains stable. See Section~\ref{sec:robust_D3BC} for details.} 
    \label{fig:sttop} 
\end{figure*}

Object detection has made significant strides in recent years. However, two advanced tasks based on this technology continue to present considerable challenges: open-vocabulary object detection (OVOD) and descriptive textual object detection, such as referring expression comprehension (REC) and visual grounding (VG). Open-vocabulary object detection aims to detect previously unseen objects in dynamic environments. Recent works 
\cite{dou2022coarse,gu2021open,li2022grounded,lin2022learning,minderer2023scaling,zhao2022exploiting,jin2024llms,zang2024contextual}
have advanced training strategies for such tasks, others 
\cite{kamath2021mdetr,kuo2022f,minderer2022simple,subramanian2022reclip}
have focused on enhancing model architectures. In parallel, tasks involving referring expressions and visual grounding, which require detecting objects based on complex natural language descriptions, have shown advances in training methodologies \cite{xie2025relationlmm,chen2025re,zong2025ground,lin2024generative,peng2023kosmos}, architectural improvements \cite{yin2025rod,lin2023sphinx,you2023ferret} and the use of the capabilities of large models \cite{shen2025vlm,xuan2024pink,zhan2024griffon}.

Despite these advancements, there is still a crucial gap in addressing the internal consistency within each input image and query. We identify an issue in existing models \cite{dou2022coarse,li2022grounded,li2023desco}: the features of the same object tend to vary significantly across different scenes, which indicates that current models may overfit to specific training backgrounds. This inconsistency not only affects the detection stability but might also degrade the model's generalization ability, raising an important question: \textit{Can we obtain object features that are robust to environmental changes?} To validate this, we construct the $D^3_{BC}$ test set by applying background replacement to the original $D^3$ dataset. Detailed in Section~\ref{sec:robust_D3BC}, baseline methods suffer notable performance drops under this setting, highlighting their limited robustness to contextual changes. In contrast, our method maintains performance comparable to that on the original benchmark. As shown in Figure~\ref{fig:sttop}, our experimental results demonstrate that addressing this issue significantly improves model performance.

To address this issue, we propose the CCL framework that enforces invariance of object features across different scenes, as shown in Figure~\ref{fig:overview}. However, existing datasets exhibit a notable limitation: they lack comprehensive data pairs that depict the same object in diverse contextual settings. This data gap is crucial because CCL requires models to encounter and learn from variations of the same object in different environments or scenarios. Without such diverse representations, models struggle to generalize under varying real-world conditions. To overcome this limitation, we introduce CBDG, which first increases the number of categories and then leverages SAM~\cite{kirillov2023segment} and the Stable Diffusion model~\cite{rombach2022high} to generate data pairs across different scenes while ensuring consistent foreground objects, thus improving both category variation and background diversity in training data.

Our experimental results demonstrate significant improvements on two challenging benchmarks, $D^3$~\cite{xie2023described} and OmniLabel~\cite{schulter2023omnilabel}, achieving +16.3 AP on OmniLabel and +14.9 AP on $D^3$. 
The proposed CBDG and CCLoss are complementary components that collectively form a robust training paradigm. Specifically, the CBDG improves feature learning through diverse scene-object compositions, while the CCLoss ensures robust feature representation across varying backgrounds. Furthermore, our approach is fundamentally model-agnostic, enabling seamless integration into a wide range of existing architectures, such as~\cite{dou2022coarse,li2022grounded}, with consistent performance gains across different frameworks.

In summary, the contributions are as follows.
\begin{itemize}[left=0pt]
\item This study identifies an issue where object features are highly susceptible to environmental changes, leading to potential overfitting and poor generalization to unseen scenarios.
\item To ensure feature robustness to context changes, we propose CCL, which enforces object consistency across backgrounds via CBDG and CCLoss.
\item Our method is simple, efficient, and model-agnostic, imposing no additional inference overhead while consistently delivering performance improvements across diverse datasets and models. Moreover, despite working with a much smaller subset of the original dataset, we achieve state-of-the-art results on two descriptive open-vocabulary detection benchmarks.
\end{itemize}

%% file: sec/2_Relatedwork.tex
\section{Related work}
\label{sec:RelatedWork}

\begin{figure*}[t] 
    \centering
    \includegraphics[width=\textwidth]{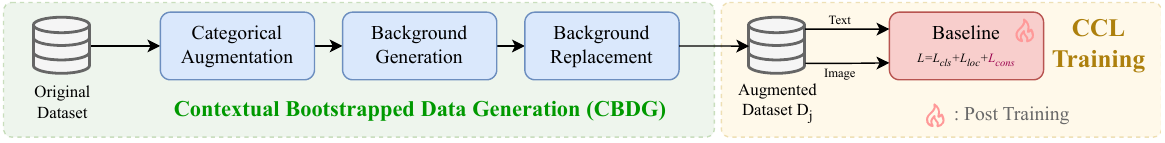} 
    \caption{Overview of our approach. CBDG generates $D_j$ via Categorical Augmentation, Background Generation and Background Replacement. CCL training uses $D_j$ with CCLoss added to total loss.} 
    \label{fig:overview} 
\end{figure*}

\paragraph{Vision language localization tasks.} Open-vocabulary object detection (OVOD) aims to enable models to recognize novel objects or unseen categories during inference
\cite{gu2021open,minderer2023scaling,zareian2021open}
, extending beyond traditional categorical detection. However, this ability is typically limited to detecting object categories based on labels, rather than understanding long descriptions. In contrast, referring expression comprehension (REC) involves understanding and localizing objects in an image based on natural language descriptions that refer to specific instances of objects
\cite{yu2016modeling,wu2020phrasecut,mao2016generation}, 
making it inherently more flexible and context-aware. While OVOD and REC both address the challenge of understanding objects in images, we focus on simultaneously handling novel categories and complex natural language descriptions. We opt for described object detection (DOD)~\cite{xie2023described} and OmniLabel~\cite{schulter2023omnilabel} as robust solutions to these challenges, as they incorporate both the recognition of novel categories and the understanding of intricate descriptions.

\paragraph{Diffusion models for scenario generation.} Stable Diffusion~\cite{rombach2022high} marks a shift in text-to-image synthesis by operating in a compressed latent space using iterative denoising. Unlike GANs~\cite{goodfellow2014generative} or VAEs~\cite{kingma2013auto} that generate images in pixel space, it uses a VAE to encode images into low-dimensional latents, allowing efficient training and high-resolution output. Guided by a pre-trained CLIP text encoder, the model aligns generated images with complex textual descriptions, from concrete objects to abstract scenes.

Recent diffusion-based methods have shown strong performance in image inpainting, enabling object and scene editing via text or spatial inputs. GLIDE~\cite{nichol2021glide} enables text-guided object replacement while preserving scene consistency, and GLIGEN~\cite{li2023gligen} extends this by incorporating bounding boxes for more precise control over object placement. For background replacement, IAM~\cite{yu2023inpaint} integrates segmentation with diffusion to regenerate regions based on textual prompts. Despite their effectiveness, these methods often suffer from boundary artifacts due to over-smoothing during denoising. We evaluate GLIDE, IAM, and Stable Diffusion~\cite{rombach2022high} in CBDG and ultimately choose Stable Diffusion for background generation.

\paragraph{Cross-modal object detection models.} With the advancement of multimodal vision language models, such as CLIP~\cite{radford2021learning} and ALIGN~\cite{jia2021scaling}, the development of methods that integrate vision and language to address visual recognition tasks has emerged as a prominent trend. GLIP~\cite{li2022grounded}, based on CLIP~\cite{radford2021learning}, leverages free-form language supervision during training and frames object detection as visual localization, constructing a foundation for semantically enriched pre-trained models. Building on this, FIBER~\cite{dou2022coarse} employs a two-stage training approach, transitioning from coarse-grained to fine-grained, enhancing the adaptability of the pre-trained model to a broad spectrum of downstream tasks at both image-level and region-level. In our work, we use GLIP~\cite{li2022grounded} and FIBER~\cite{dou2022coarse} as baseline models and incorporate our CCL method to validate the experimental results.

%% file: sec/3_Method.tex
\section{Method}
\label{sec:Method}
\subsection{Overview}
We introduce CCL, a novel framework designed to address the challenge of maintaining detection and grounding consistency when models encounter diverse and unseen object categories across varying contextual backgrounds. To achieve this goal, we address two fundamental aspects of the problem: the lack of appropriate training data and the need for effective consistency-preserving mechanisms.


In Section~\ref{sec:data_augmentation }, we describe our CBDG pipeline, which leverages advanced segmentation and generative models to create a rich and varied dataset. This data preparation process is specifically designed to support our consistency learning objectives. Following this, in Section~\ref{sec:consistency }, we detail our CCLoss formulation, which ensures that the model learns to maintain object identity across different backgrounds. 

CBDG and CCLoss serve complementary but distinct roles. CBDG is not a generic augmentation module, instead, it constructs identity-preserving paired samples in which the same object appears under different contextual backgrounds. These pairs define the supervision signal for consistency learning, but do not enforce invariance by themselves. CCLoss is the objective that ties these paired views and drives the model to learn background-invariant representations. In this sense, CBDG provides the data structure, while CCLoss converts it into consistency-aware learning.

\subsection{Contextual Bootstrapped Data Generation}
\label{sec:data_augmentation }
\begin{figure*}[t] 
    \centering
    \includegraphics[width=\textwidth]{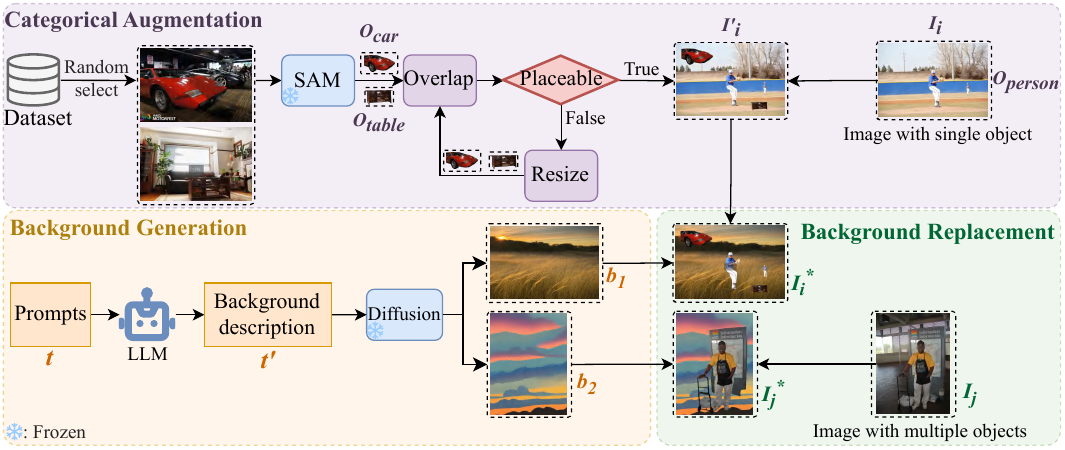} 
    \caption{CBDG Pipeline. We use ChatGPT to generate background prompts for a diffusion model, enabling diverse background synthesis. For single-class images, CBDG augments object categories before background replacement. For multi-class images, CBDG replaces only the background.} 
    \label{fig:dataaug} 
\end{figure*}

Current open-set visual grounding methods struggle to maintain robustness across diverse real-world scenarios, particularly when objects appear in unfamiliar contextual settings. This limitation stems from a fundamental data scarcity: existing datasets rarely capture the full spectrum of object-background interactions, leading to biased model performance. To overcome this, we propose a multistage data augmentation framework that synthesizes diverse and realistic object-context compositions by combining SAM-based object manipulation with text-guided background generation, as shown in Figure~\ref{fig:dataaug}. Our method constructs a compositionally diverse joint dataset $D_j$ that mitigates common inpainting artifacts and improves model generalization.

\paragraph{Categorical augmentation.} In our approach, we use the Flickr30k Entities visual grounding dataset~\cite{plummer2015flickr30k} alongside a subset of the Objects365 object detection dataset~\cite{shao2019objects365} to create a combined training dataset. The selected subset of Objects365 tends to contain images dominated by a few categories, with multiple instances of that category present. Details are discussed in Supplementary Section~\ref{PPDD} and Section~\ref{ISandCA}.


For images with a single object, we aim to enhance the diversity of object categories within each image by introducing objects from different categories while maintaining spatial and contextual coherence. To achieve this, we leverage the SAM model~\cite{kirillov2023segment} to extract precise objects $O_i$ and the corresponding position $(x_o, y_o)$ for each image $I$, where $i$ means the category ID to which the object belongs. The object masks allow us to identify individual objects and their spatial locations. Based on these masks, we randomly select objects $O_{i\notin C}$ from other images within the same subset but belonging to different categories, where $C$ represents the category set. Then these objects are positioned in the current image at carefully chosen locations. Specifically, we define $P = \{(x_1,y_1), (x_2,y_2), ..., (x_N,y_N)\}$ the potential placement position set for the new object, $N$ is the number of candidate locations. From these positions, we randomly select $(x, y)\in P\backslash(x_o,y_o)$ that does not overlap with the existing objects in the image, ensuring a clean and non-interfering insertion of the new object. This process of placement can be formalized as:
\begin{equation}
    \texttt{Augmentation}: (x_{o_k}, y_{o_k})\in P\backslash(x_o,y_o), o_k \in O_{i\notin C},
    \label{eq:categ}
\end{equation}
where $k$ represents the category of selected object, $(x_{o_k}, y_{o_k})$ denotes the position randomly chosen from the set of candidate locations according to the above rule. After categorical augmentation, the original image $I$ becomes $I'$.

In scenarios where no suitable empty position is available, such as when the current image contains large objects or a large number of dispersed objects, which results in limited available space, we adopt a resizing strategy. In these cases, we reduce the size of the new object to $ 1/\alpha $ of its original size and attempt to place it again, where $\alpha$ is a scaling factor. This process is repeated until an empty placement area is found or the number of resizing attempts exceeds a threshold $ N_R $. If resizing attempts fail to find a suitable location, we abandon the current image and instead select another image to enhance the diversity of object categories, thus ensuring a broader range of category representation in the final dataset.

\paragraph{Background generation.} With more object categories added, the foreground dataset now includes images with varied labels and their corresponding bounding boxes. To reduce model overfitting and improve generalization, we next generate diverse background images, placing the same objects in different scenes. Image inpainting methods \cite{nichol2021glide, li2023gligen, yu2023inpaint} often struggle with blurred edges and backgrounds that still reflect foreground features, making realistic scene changes difficult (see Supplementary Section~\ref{IMBC}). To avoid these issues, we use a simpler alternative that better separates foreground from background.

Instead of relying on limited original image content, we generate new and simple backgrounds directly. Using a Large Language Model (LLM)~\cite{brown2020language}, denoted as $\mathcal{G}$, we create text prompts in three categories: \textit{Seasonal, Sky, and Natural Landscape}, to ensure variety and relevance. Details of these prompts are provided in Supplementary Section~\ref{promptLLM}. These prompts are input into Stable Diffusion $\mathcal{D}$~\cite{rombach2022high}, which generates matching background images. This method allows us to build a diverse, context-aware background dataset without the limitations of inpainting:
\begin{equation}
    \texttt{Generation}:b=\mathcal{D}(t'), t' = \mathcal{G}(t),
    \label{eq:diffbg}
\end{equation}
where $t \in $ \{\textit{Seasonal, Sky, Natural Landscape}\}, $t'$ is the background description generated by ChatGPT, and $b$ represents the background image generated by stable Diffusion, which constitutes the background dataset $D_{bg}$. We start with visually clean and semantically neutral backgrounds to isolate the effect of contextual variation on object-level consistency. Beyond the three primary prompt families, we further analyze urban/indoor/architectural scenes in Supplementary Section~\ref{SceneDiversity} and Section~\ref{ABG}.

\begin{figure}[t] 
    \centering
    \includegraphics[width=\columnwidth]{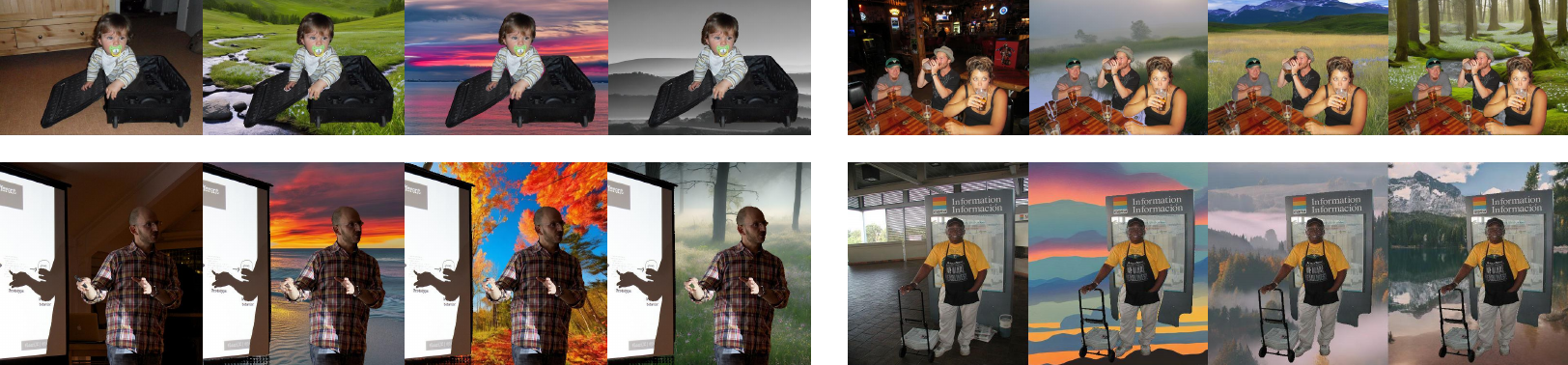} 
    \caption{Four groups of images are shown, each composed of four sub-images: the leftmost sub-image in every group is the original, while the remaining three display background replacements.} 
    \label{fig:fourimages} 
\end{figure}


\paragraph{Background replacement.} At this stage, we have the generated background dataset $D_{bg}$ and the foreground dataset $D_{fg}$ covering various object categories. To construct new contextual variations, we randomly pair each foreground image with background images sampled from $D_{bg}$. Given the ground-truth bounding boxes, we use the SAM model $\mathcal{S}$~\cite{kirillov2023segment} to extract foreground objects. To ensure compositing quality, we filter out samples whose SAM-derived mask produces a bounding box with an IoU below a predefined threshold with respect to the ground-truth box. Implementation details are provided in Supplementary Section~\ref{EITTQ}. After isolating the foreground, we replace the original background with a selected one, generating multiple new images per original. The foreground stays the same, while the backgrounds vary, producing diverse scenes with consistent object content. The replacement process is defined as:
\begin{equation}
    \texttt{Replacement}:I^*=\mathcal{S}(I', bbox)\oplus b, b\in D_{bg},
    \label{eq:backrep}
\end{equation}
where $bbox$ represents the bounding boxes of objects in the image $I'$ after categorical augmentation, $\oplus$ denotes the composition of foreground and background, $I^*$ represents the image with replaced background. 

CBDG enables us to significantly augment the dataset with diverse background settings while maintaining the integrity of the foreground objects, providing a more robust foundation for training our model. As shown in Figure~\ref{fig:fourimages}, after CBDG, for each original image, several additional images are generated with replaced backgrounds. This results in a total of $K$ images per original, all sharing the same foreground objects, but differing in their backgrounds. $K$ represents the batch size used during training. Alternative data generation schemes are also compared, see Supplementary Section~\ref{GenerationScheme}. The augmented images are then utilized for the subsequent consistency constraints in our approach.

\begin{figure*}[t]  
    \centering
    \includegraphics[width=\textwidth]{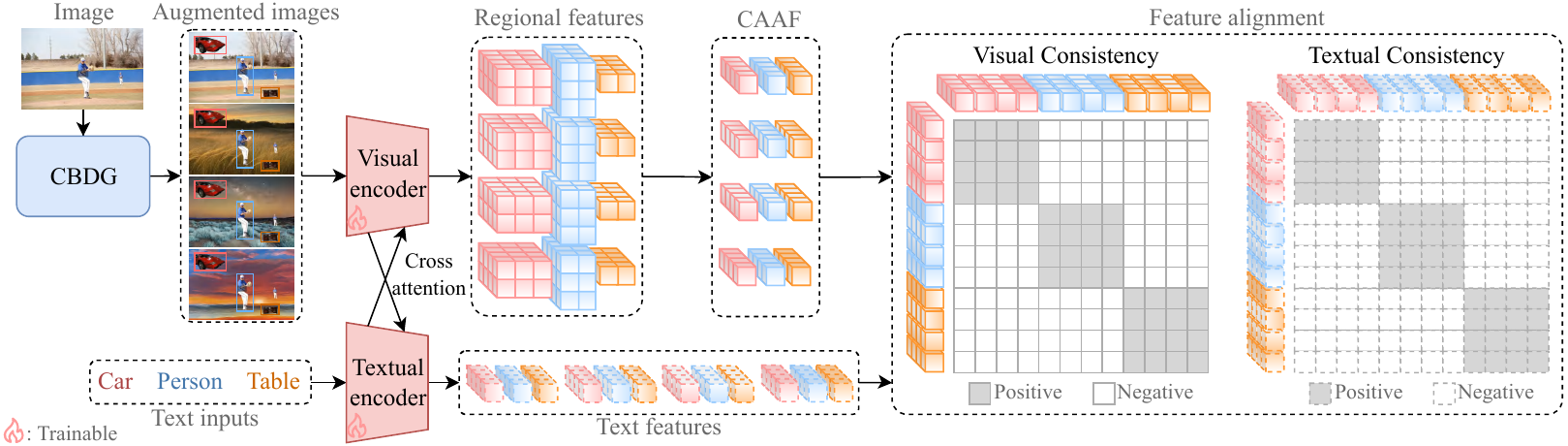}  
    \caption{CCL Framework. Visual and textual features are encoded, with regional features pooled into CAAF. Consistency loss is applied within each modality.}
    \label{fig:pipeline}
\end{figure*}

\subsection{Contextual Consistency Loss}
\label{sec:consistency }
Given that after CBDG, we now have access to a dataset $D_j$ in which each group of images shares the same foreground object but varies in background. We introduce the Contextual Consistency Loss (CCLoss), a novel training objective designed to enforce representation invariance for the same object category across varying contextual environments. As illustrated in Figure~\ref{fig:pipeline}, our method uses CCLoss to maintain the consistency of foreground object representations across different backgrounds. By constructing training batches that contain instances of the same object under different contextual settings, CCLoss encourages the model to focus on semantically meaningful foreground features rather than background-dependent or spurious cues. This section elaborates on the underlying model architecture, the detailed formulation of consistency loss, and its integration into the overall training objective.

\paragraph{Model architecture.} 
We employ a language-based object detector~\cite{li2022grounded,dou2022coarse} as the backbone for feature extraction and object detection, taking advantage of its strong capability in bridging vision and language representations. Specifically, images and textual descriptions are first encoded to obtain their respective feature embeddings, ensuring a comprehensive understanding of both modalities. These extracted features are subsequently processed through a Feature Pyramid Network (FPN), which effectively refines and integrates multiscale representations, thereby enhancing detection performance across various object sizes and contexts. To further improve localization accuracy, the refined image features are then passed on to DynamicHead, a dedicated module designed to predict a set of candidate regions where objects are most likely to be located. This hierarchical and adaptive processing pipeline ensures robust and efficient object detection.

\paragraph{Consistency loss.} 
During the training phase, we organize each batch by grouping images that share identical foreground objects but exhibit diverse background settings. This arrangement enables the computation of the CCLoss function, which serves as a critical mechanism for training the model to preserve invariant representations of foreground objects across varying contextual environments.

The total loss function, as depicted in Eq.~\ref{eq:1}, comprises three fundamental components: localization loss, classification loss, and contextual consistency loss ($\mathcal{L}_{\mathrm{cons}}$). Each of these components contributes to optimizing the model performance in different aspects: precise object localization, accurate category classification, and robust feature representation that maintains foreground consistency irrespective of background variations. The first two components of the loss function are detailed in GLIP~\cite{li2022grounded}. The integration of these loss terms ensures a balanced optimization process that addresses discriminative and invariant feature learning.
\begin{equation}
    \mathcal{L} = \mathcal{L}_{\mathrm{cls}} + \mathcal{L}_{\mathrm{loc}} + \mathcal{L}_{\mathrm{cons}},
    \label{eq:1}
\end{equation}
Eq.~\ref{eq:2} provides the formulation of CCLoss. CCLoss combines the text and image modality losses with weighting factors.  $\lambda_{\mathrm{T}}$ and $\lambda_{\mathrm{I}}$ are weighting parameters to balance the loss contributions in the text and the image modality.
\begin{equation}
        \mathcal{L}_{\mathrm{cons}} = \lambda_{\mathrm{T}} \cdot \mathcal{L}_{\mathrm{T}} + \lambda_{\mathrm{I}} \cdot \mathcal{L}_{\mathrm{I}},
    \label{eq:2}
\end{equation}
For image features obtained from the image encoder, we first perform a pooling operation on them to obtain the Context-Aware Aggregated Feature (CAAF), denoted as $f$, followed by applying a consistency loss among the CAAF. Given a batch with $C$ categories and $K$ images, the contrastive loss for the vision modality is defined as:
\begin{equation}
    \mathcal{L}_{\mathrm{I}} = -\frac{1}{C K} \sum_{c=1}^{C} \sum_{k=1}^{K} \log \frac{\exp\bigl(\mathrm{sim}(\mathbf{f}_{ck}, \mathbf{f}_{c}) / \tau\bigr)}{\sum\limits_{c'=1}^{C} \sum\limits_{k'=1}^{K} \exp\bigl(\mathrm{sim}(\mathbf{f}_{ck}, \mathbf{f}_{c'k'}) / \tau\bigr)},
    \label{eq:3}
\end{equation}
where $\mathbf{f}_{ck}$ is the $k$-th image feature of the $c$-th category. $\mathbf{f}_{c'k'}$ is the $k'$-th image feature of the $c'$-th category. $\mathbf{f}_{c}$ is the centroid of the image features for the $c$-th category, calculated as the mean of the $K$ image features. $\mathrm{sim}(\cdot, \cdot)$ is cosine similarity. $\tau$ is the temperature parameter.

Similarly, for text features, we implement a contrastive learning objective that promotes feature clustering within the same category while enforcing separation among different categories. However, the application of this text contrastive loss is contingent upon the baseline architecture: When using FIBER as the baseline, where cross-modal interactions between image and text encoders are enabled, we fully utilize this loss term. In contrast, when employing GLIP as the baseline, which processes image and text modalities independently, we effectively disable this component by setting its weight $\lambda_{\text{T}}$ to zero. Given a batch with $C$ categories and $K$ images, the contrastive loss for the text modality is defined as:
\begin{equation}
    {
        \mathcal{L}_{\mathrm{T}} = -\frac{1}{C K} \sum_{c=1}^{C} \sum_{k=1}^{K} \log \frac{\exp\bigl(\mathrm{sim}(\mathbf{t}_{ck}, \mathbf{t}_{c}) / \tau\bigr)}{\sum\limits_{c'=1}^{C} \sum\limits_{k'=1}^{K} \exp\bigl(\mathrm{sim}(\mathbf{t}_{ck}, \mathbf{t}_{c'k'}) / \tau\bigr)},
    }
    \label{eq:4}
\end{equation}

where $\mathbf{t}_{ck}$ is the $k$-th text feature of the $c$-th category. $\mathbf{t}_{c'k'}$ is the $k'$-th text feature of the $c'$-th category. $\mathbf{t}_{c}$ is the centroid of the text features for the $c$-th category, calculated as the mean of the $K$ text features. The design of our CCLoss follows a progressive evolution, with the detailed process provided in Supplementary Section~\ref{PDICL}.

%% file: sec/4_experiment.tex
\section{Experiments}
\label{sec:Experiments}

\begin{table*}[t] 
\caption{Performance of our method compared with SOTA methods.}
\centering
\resizebox{\textwidth}{!}{
\begin{tabular}{c|ccccccc|ccc}
\toprule
 & \multicolumn{7}{c|}{OmniLabel} & \multicolumn{3}{c}{$D^{3}$} \\ 
\cmidrule(lr){2-8} \cmidrule(lr){9-11}
Method & AP & AP-c & AP-d & AP-dP & AP-dS & AP-dM & AP-dL & FULL & PRES & ABS \\
\midrule
Detic~\cite{zhou2022detecting} & 8.0 & 15.6 & 5.4 & 8.0 & 5.7 & 5.4 & 6.2 & - & - & -\\
OFA-DOD~\cite{xie2023described} & - & - & - & - & - & - & - & 21.6 & 23.7 & 15.4\\
RelationLLM-L~\cite{xie2025relationlmm} & - & - & - & - & - & - & - & 24.3 & 24.6 & 23.4\\
GN-GLIP~\cite{zhao2024generating} & 22.2 & 27.2 & 18.8 & 29.0 & - & - & - & 21.4 & 20.6 & 23.7\\
GN-FIBER~\cite{zhao2024generating} & 28.1 & 32.1 & 25.1 & 36.5 & - & - & - & 26.0 & 25.2 & 28.1\\
ROD-MLLM~\cite{yin2025rod} & - & - & 25.3 & 30.9 & 31.8 & 24.5 & 21.0 & 29.7 & 30.0 & 28.7\\
Real-Model~\cite{chen2025re} & - & - & 36.5 & \textbf{52.1} & \textbf{54.4} & 33.2 & 25.5 & 34.1 & 34.4 & 33.2\\

\midrule
GLIP-T~\cite{li2022grounded} & 19.3 & 23.6 & 16.4 & 25.8 & 29.4 & 14.8 & 8.2 & 19.1 & 18.3 & 21.5\\
+ours & 32.2 & 36.1 & 28.8 & 39.8 & 43.3 & 26.5 & 17.6 & 30.0 & 29.2 & 32.3\\

\midrule
FIBER-B~\cite{dou2022coarse} & 25.7 & 30.3 & 22.3 & 34.8 & 38.6 & 19.5 & 12.4 & 22.7 & 21.5 & 26.0\\
+ours & \textbf{42.0} & \textbf{44.1} & \textbf{39.2} & 50.8 & 53.7 & \textbf{38.2} & \textbf{32.3} & \textbf{37.6} & \textbf{37.2} & \textbf{38.8}\\

\bottomrule
\end{tabular}
}
\label{tabt1}
\end{table*}

\subsection{Experimental Design}

\paragraph{Training setup.} To evaluate the generalizability of our proposed method, we use two baseline models, GLIP~\cite{li2022grounded} and FIBER~\cite{dou2022coarse}. These models serve as benchmarks for comparison. The datasets used to train the baseline models are 1) Objects365 (O365)~\cite{shao2019objects365} and 2) GoldG, including Flickr30K~\cite{plummer2015flickr30k}, VG Caption~\cite{krishna2017visual}, and GQA~\cite{hudson2019gqa}, which together contain 0.8 million images, providing a diverse and large-scale training set.

In contrast, for our method, we work with a smaller subset of the original dataset, with only 0.25 million images as the initial joint dataset for CBDG. Specifically, we incorporate the Flickr30k Entities~\cite{plummer2015flickr30k} dataset along with only 0.22M images of the Objects365 dataset~\cite{shao2019objects365}, which is much smaller than the full dataset used for the baselines. 

We generate three main categories of background images in CBDG: seasonal, sky, and natural landscape. In total, we have 13,185 unique descriptions, resulting in 144,654 generated images. The breakdown of categories and the corresponding number of images is as follows: seasonal (3387 descriptions, 48,156 images), sky (3399 descriptions, 48,210 images), and natural landscape (3399 descriptions, 48,288 images).

For training, we use publicly available pre-trained model checkpoints of both GLIP~\cite{li2022grounded} and FIBER~\cite{dou2022coarse}. These pre-trained weights serve as the starting point for fine-tuning. We fine-tune the model for one epoch on our dataset $D_j$. After this fine-tuning process, we obtain the final results, which demonstrate the effectiveness of our method when applied to a smaller and more constrained dataset. The implementation details and computational cost can be found in Supplementary Section~\ref{details}. We report the choice and tuning of hyperparameters in Supplementary Section~\ref{Hyperparameters}.

\paragraph{Benchmark selection.} We choose OmniLabel~\cite{schulter2023omnilabel} and $D^3$~\cite{xie2023described} as benchmark evaluation methods, both of which use Average Precision (AP) as the evaluation metric. The reason we select these two benchmarks is that they not only provide object category labels but also include a rich diversity of textual descriptions, which place a greater emphasis on the model’s ability to understand and interpret language. This aspect makes these benchmarks particularly valuable for evaluating the model’s performance in tasks involving both visual and linguistic information. Compared to other REC 
\cite{yu2016modeling,wu2020phrasecut,mao2016generation}
and OVOD 
\cite{gupta2019lvis}
benchmarks, $D^3$~\cite{xie2023described} and OmniLabel~\cite{schulter2023omnilabel} offer a broader evaluation of object detection capabilities. These benchmarks include negative samples and more precisely defined bounding boxes corresponding to textual descriptions, which can refer to zero, one, or multiple objects in the image. This makes the tasks more challenging and forces the model to effectively localize and recognize objects based on a range of different descriptions and contexts, offering a more comprehensive test of its generalization and performance in diverse scenarios.

\subsection{Comparison with SOTA Methods}
Table~\ref{tabt1} presents a comparison between our method and the current SOTA methods on the OmniLabel~\cite{schulter2023omnilabel} and $D^3$~\cite{xie2023described} benchmarks. The first column lists various model methods, followed by seven columns representing the seven AP metrics on OmniLabel. These metrics include: plain categories (AP-c) and free-form descriptions (AP-d). AP-dP evaluates only positive descriptions. AP-dS/M/L assess descriptions of varying lengths (up to 3 words, 4-8 words, and more than 8 words). The last three columns represent the AP metrics on $D^3$: FULL, PRES, and ABS, which evaluate all descriptions, only presence descriptions, and only absence descriptions, respectively. 

We use GLIP-T~\cite{li2022grounded} and FIBER-B~\cite{dou2022coarse} as baselines and fine-tune them on our method. With the integration of our proposed CCL method, significant improvements are observed across multiple benchmarks. Specifically, when applied to the FIBER baseline, the method achieves a notable increase of +16.3 AP on the OmniLabel benchmark and +14.9 AP on the $D^3$ benchmark. Similarly, when implemented with the GLIP baseline, our method demonstrates consistent performance gains, achieving +12.9 AP on the OmniLabel benchmark and +10.9 AP on the $D^3$ benchmark. These results underscore the effectiveness of our approach in improving contextual understanding and consistency across diverse datasets. We further evaluate our method on phrase grounding tasks to demonstrate broader applicability (see Supplementary Section~\ref{discussiona}).


\subsection{Robustness Evaluation under Background Variations}
\label{sec:robust_D3BC}

To quantitatively assess the robustness of OVOD models under contextual and domain variations, we conduct two complementary evaluations based on the $D^3$ benchmark. First, to isolate the effect of background shifts, we construct a new benchmark derived from $D^3$. For each of the 10,578 original images in $D^3$, we generate three additional variants by replacing the background using the CBDG pipeline proposed in this work. These new background images are generated independently of the training data, ensuring no overlap or information leakage. The resulting dataset, termed $D^3_{\mathrm{BC}}$, consists of the original images and their background-altered counterparts, totaling 42,312 samples. We evaluate two representative baseline models, GLIP-T and FIBER-B, on both $D^3$ and $D^3_{\mathrm{BC}}$, and further examine their performance when enhanced with our proposed CCL method. This yields four experimental settings. As summarized in Table~\ref{tab:t4}, both baselines exhibit substantial performance degradation on $D^3_{\mathrm{BC}}$, revealing their susceptibility to background shifts. However, models incorporating our CCL approach demonstrate significantly improved robustness with small performance drops, demonstrating that our method improves robustness to contextual background changes.

\begin{figure}[t] 
    \centering
    \includegraphics[width=\columnwidth]{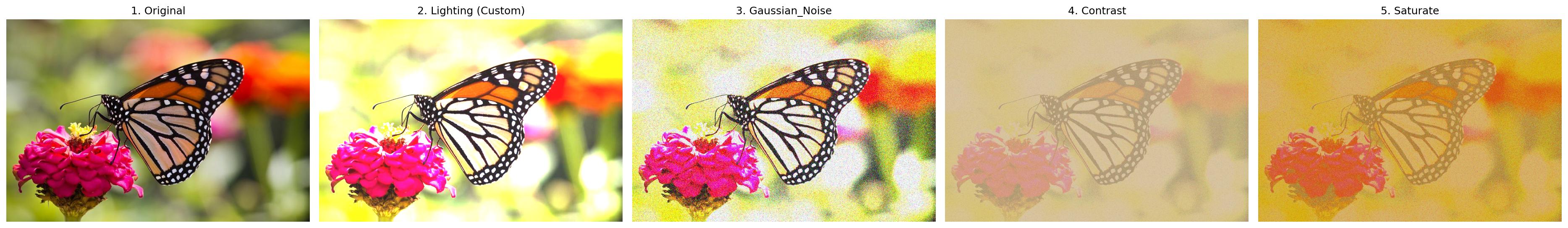} 
    \caption{Examples of $D^3_{\mathrm{C}}$ Perturbations.} 
    \label{fig:D3C} 
\end{figure}

To further verify that CBDG + CCL does not increase sensitivity to non-contextual domain shifts, we additionally construct $D^3_{\mathrm{C}}$ by following the COCO-C~\cite{michaelis2019benchmarking} protocol and applying four perturbations to the original $D^3$ images, including Gaussian Noise, Contrast, Saturation, and Lighting. Following prior robustness evaluation practice, we report both the mean corrupted performance and the relative robustness: 
\begin{equation}
{
m\mathrm{FULL} = \frac{1}{K}\sum_{k=1}^{K}\mathrm{FULL}_k,\quad
r\mathrm{FULL} = \frac{m\mathrm{FULL}}{\mathrm{FULL}} \times 100\%
}, 
\label{mfull}
\end{equation}
where $K$ denotes the number of corruption types. As shown in Table~\ref{tab:t4}, CCL achieves substantially higher absolute performance than the corresponding baselines on $D^3_{\mathrm{C}}$, while maintaining comparable relative degradation, indicating background robustness is improved without sacrificing stability under other domain shifts.

Taken together, these results show that CCL not only improves robustness under explicit background variations, but also preserves stability under other common domain shifts, indicating that the gains are not limited to the specific contextual perturbations introduced by CBDG.

\subsection{Ablation Study on CBDG and CCLoss}
Given that our method is fundamentally grounded in consistency and incorporates a certain degree of data generation, we perform a series of ablation experiments to evaluate the contribution of each individual component. In particular, we conduct two distinct experimental setups to assess the impact of both CBDG and CCLoss. The first experiment introduces CBDG to the baseline model, followed by fine-tuning the model for one epoch on $D_j$. To ensure a fair comparison, we keep the training parameters consistent with those used in the baseline experiment. The second experiment represents our complete experimental setup, adding CBDG and CCLoss to the baseline model and fine-tuning the model for one epoch. As shown in Table~\ref{tab:t2}, both CBDG and CCLoss play an essential role in enhancing the model’s performance. CBDG increases the diversity of training data, improving the model’s robustness across varying conditions. Meanwhile, the CCLoss reinforces object consistency across different contexts, ensuring that the model can reliably detect and localize objects regardless of their surrounding environment. The combined effects of these two components contribute significantly to the observed performance improvements. We further analyze the impact of dataset scale in Section~\ref{datascale}.

\begin{table}[t]
\centering
\caption{Performance comparison on {$D^3$\textsubscript{BC}} and {$D^3_{\mathrm{C}}$}.}
\resizebox{\linewidth}{!}{ 
\begin{tabular}{c|ccc|ccc}
  \toprule
  & \multicolumn{3}{c|}{$D^3$\textsubscript{BC}} & \multicolumn{3}{c}{$D^3_{\mathrm{C}}$}\\
  Method & FULL & PRES & ABS & FULL & mFULL & rFULL\\
  \midrule
  GLIP-T & 16.8 & 15.8 & 20.4 & 19.1 & 13.6 & 71.2\\
  +ours  & 29.6 & 28.9 & 31.9 & 30.0 & 21.7 & 72.3\\
  \midrule
  FIBER-B & 20.1 & 18.7 & 24.7 & 22.7 & 16.7 & 73.6\\
  +ours   & 33.1 & 32.8 & 34.0 & 37.6 & 27.5 & 73.1\\
  \bottomrule
\end{tabular}
}
\label{tab:t4}
\end{table}

\begin{table}[t]
\centering
\caption{Ablation study on the effects of Contextual Bootstrapped Data Generation (CBDG) and Contextual Consistency Loss (CCLoss). “+data” denotes using CBDG without CCLoss, “+ours” combines both components.}
\resizebox{\linewidth}{!}{ 
\begin{tabular}{c|ccc|ccc}
\toprule
 & \multicolumn{3}{c|}{OmniLabel} & \multicolumn{3}{c}{$D^{3}$} \\
Method & AP & AP-c & AP-d & FULL & PRES & ABS \\
\midrule
GLIP-T  & 19.3 & 23.6 & 16.4 & 19.1 & 18.3 & 21.5\\
+data   & 24.8 & 29.2 & 21.8 & 23.2 & 22.5 & 25.3\\
+ours   & 32.2 & 36.1 & 28.8 & 30.0 & 29.2 & 32.3\\
\midrule
FIBER-B & 25.7 & 30.3 & 22.3 & 22.7 & 21.5 & 26.0\\
+data   & 32.7 & 35.8 & 29.6 & 29.1 & 28.3 & 31.2\\
+ours   & 42.0 & 44.1 & 39.2 & 37.6 & 37.2 & 38.8\\
\bottomrule
\end{tabular}
}
\label{tab:t2}
\end{table}

%% file: sec/5_conclusion.tex
\section{Conclusion}
\label{sec:Conclusion}





\paragraph{Summary.} We propose CCL, a simple and effective framework for descriptive open-vocabulary object detection. By combining CBDG for contextual data generation and CCLoss for feature consistency learning, CCL improves generalization under contextual variation with less data and no extra inference cost. The framework is model-agnostic, easy to integrate, yielding consistent gains across benchmarks.

\paragraph{Limitation \& Future Work.} Our method relies on the quality of SAM-based segmentation. Imperfect masks can introduce artifacts during CBDG. Although post-processing alleviates these effects, developing more artifact-free generation pipelines remains important for future work.

%% file: sec/ack.tex
\paragraph{Acknowledgments}\mbox{}\\
This work was supported by the project of Pengcheng Laboratory (PCL2025A14), by National Natural Science Foundation of China (grant No. 62350710797), by Guangdong Basic and Applied Basic Research Foundation (grant No. 2023B1515120065, 2025A1515011546) and by the Shenzhen Science and Technology Program
(JCYJ20240813105901003, KJZD20240903102901003, ZDCY20250901113000001).

%% file: sec/X_suppl.tex
\clearpage
\setcounter{page}{1}
\appendix

\twocolumn[
\begin{center}
    {\LARGE \textbf{Consistency Beyond Contrast: Enhancing Open-Vocabulary Object Detection Robustness via Contextual Consistency Learning}\par}
    \vspace{0.4em}
    {\LARGE Supplementary Material\par}
    \vspace{1.2em}
\end{center}
]

This document supplements the main paper as follows.

{A. Details}
\begin{itemize}
  \item Sect.~\ref{Ids} presents the implementation details and computational cost.
\end{itemize}

{B. Additional Experiments}
\begin{itemize}
  \item Sect.~\ref{Hyperparameters} presents the choice of hyperparameters.
  \item Sect.~\ref{GenerationScheme} presents the choice of generation scheme.
  \item Sect.~\ref{SceneDiversity} presents the analysis on CBDG scene diversity.
  \item Sect.~\ref{discussiona} presents additional experiments on phrase grounding.
  \item Sect.~\ref{IMBC} presents an analysis of inpainting methods for background change.
  \item Sect.~\ref{PDICL} presents the progressive design of the intra-modal consistency loss.
  \item Sect.~\ref{datascale} presents the analysis on data scale.
\end{itemize}

{C. Related Work}
\begin{itemize}
  \item Sect.~\ref{ROD} presents the related work of robust object detection.
  \item Sect.~\ref{DAD} presents the related work of data augmentation for detectors.
  \item Sect.~\ref{CMODM} presents the related work of cross-modal object detection models.
  \item Sect.~\ref{existingdata} presents the survey of existing background datasets.
\end{itemize}

{D. CBDG Further Analysis}
\begin{itemize}
  \item Sect.~\ref{semanticD} discusses the semantic conflicts and object-background decoupling.
  \item Sect.~\ref{promptLLM} presents the prompts used for background generation with LLM.
  \item Sect.~\ref{EITTQ} presents the post-processing techniques during background replacement stage of CBDG.
  \item Sect.~\ref{PPDD} provides an analysis of the processing procedures for different datasets.
  \item Sect.~\ref{ABG} presents the analysis on background generation.
  \item Sect.~\ref{ISandCA} provides an analysis of image selection and category augmentation.
  \item Sect.~\ref{pseudo code} presents the pseudocode of the CBDG pipeline.
\end{itemize}

\section{Details}
\label{details}

\subsection{Implementation Details and Computational Cost}
\label{Ids}

Our CCL framework consists of two main components: CBDG and training with CCLoss. The experimental details are as follows:

For Categories Augmentation, we select $P$ potential object placement locations, setting the number of positions $N=100$. If the number of available placement locations is fewer than 5, the object's width and height are scaled down by a factor of $ 1/\alpha $, where $ \alpha=2$. If the number of scaling operations exceeds a threshold $ N_R $ , we switch to another object, with $ N_R=2 $.

For model fine-tuning, we initialize GLIP and FIBER with their official checkpoints, which are trained on large-scale and diverse datasets such as GoldG and Objects365. We then fine-tune these models on our augmented dataset, generated via the CBDG pipeline. To ensure the models can adapt effectively to our domain-shifted data, we reset the learning rate to 1e-4 and use the Adam optimizer. This relatively high learning rate facilitates optimization toward new minima that are better aligned with our objective of contextual robustness, rather than preserving the prior distribution learned from the baseline training set. Both baseline models are fine-tuned with a batch size of 4, while all other training hyperparameters remain unchanged. The weight settings for CCLoss during training are as follows: When using FIBER as the baseline, $\lambda_{\text{T}}$ is set to 0.05, and $\lambda_{\text{I}}$ is set to 0.15. When using GLIP as the baseline, $\lambda_{\text{T}}$ is set to 0, and $\lambda_{\text{I}}$ is set to 0.15. The temperature parameter $\tau$ in the loss function is set to 1 by default.

While our method introduces an additional data generation step (CBDG), it is important to note that our overall training efficiency remains significantly higher than that of baseline methods. Within the CBDG stage, the most time-consuming part is the background generation step. We generated a total of 144,654 unique background images, which takes about 101 hours on 4 A100 GPUs. However, this step is conducted only once, and the generated data is reused across models and experiments. Other steps in CBDG (e.g., foreground segmentation and compositing) are relatively fast and negligible in cost.

The CCL post-training stage is lightweight. On 4 NVIDIA A100-SXM4-40G GPUs, it takes approximately 7 hours for FIBER and 9 hours for GLIP. This is significantly lower than the 540 GPU-hours reported for FIBER's original fine-grained pretraining. In our case, we only perform a single-epoch fine-tuning, which highlights the data efficiency enabled by our framework.

The following results are measured with an input image resolution of 224×224. When using GLIP as the baseline, the model has 54.81 GFLOPs and 195.19M parameters. When using FIBER as the baseline, it has 51.98 GFLOPs and 200.02M parameters. In both cases, our method introduces less than 1\% additional FLOPs and parameters relative to the baseline, resulting in negligible overhead. This demonstrates that our framework preserves computational efficiency while delivering consistent performance gains.

\section{Experiment}

\subsection{Choices of Hyperparameters}
\label{Hyperparameters}

In our experiments, we perform systematic tuning on the FIBER baseline to optimize the performance of the proposed CCLoss. Our tuning procedure is conducted in two stages. First, we fix the temperature parameter $\tau$ to its commonly used default value of 1.0, following prior contrastive learning literature, such as in CLIP and SimCLR. This choice has been widely validated to provide a good balance between hard and easy negatives without introducing instability in the optimization process. Moreover, we find that performance is relatively insensitive to minor variations in $\tau$, so we opt to retain the default value and focus our tuning efforts on the loss weight parameters.

As shown in Table~\ref{tabtuning}, we begin by tuning $\lambda_I$, the weight of loss of consistency on the image side. We vary this parameter while holding $\lambda_T$ = $\lambda_I$, and find that a value around 0.15 yields the best trade-off between stability and performance gain. Once $\lambda_I$ is fixed to 0.15, we then tune $\lambda_T$, the weight of loss of consistency on the text side. A value near 0.05 proves to be most effective when applied with the FIBER baseline, which supports cross-modal interactions. Note that for the GLIP baseline, we set the same $\lambda_I=0.15$ and set $\lambda_T$=0 since GLIP decouples visual and textual modalities, rendering this loss component ineffective in that context.

\begin{table}[ht]
\centering
\caption{Tuning on the FIBER baseline}
\begin{tabular}{c c c c c}
\toprule
Tuning & $\lambda_I$ & $\lambda_T$ & OmniLabel (AP) & $D^3$ (FULL) \\
\midrule
0 & 0   & 0    & 32.7 & 29.2 \\
1 & 0.5 & 0.5  & 35.8 & 31.5 \\
2 & 0.2 & 0.2  & 38.2 & 34.0 \\
3 & 0.1 & 0.1  & 39.9 & 36.1 \\
4 & 0.15 & 0.15  & 41.2 & 37.1 \\
5 & 0.15 & 0.1  & 41.7 & 37.4 \\
6 & 0.15 & 0.05  & 42.0 & 37.6 \\

\bottomrule
\end{tabular}
\label{tabtuning}
\end{table}

\subsection{Choice of Generation Scheme}
\label{GenerationScheme}
\paragraph{Additional Experiment on BBox Copy-Paste Method.} To further explain our choice of generation scheme, we conduct an additional study comparing our pipeline to a standard copy-paste baseline. In this experiment, we use FIBER-B as the backbone and retain the same training schedule, and hyperparameters. The only difference lies in how foreground objects are extracted: instead of using SAM-generated masks, we directly use the bounding boxes from the training set and paste the entire box region onto new backgrounds. The results are shown in Table~\ref{copypaste}, in comparison with FIBER-B+ours from Table~\ref{tabt1} of the main paper.

As seen in Table~\ref{copypaste}, the simple bbox copy-paste approach performs significantly worse than our full method. Although it does decouple foreground and background to some extent, the pasted bbox regions inevitably contain substantial non-foreground pixels, leading to impure representations. This degrades the model’s ability to learn true background-invariant features—even with the added consistency loss—since the model tends to focus on the entire bbox region rather than the precise shape of the object.

\paragraph{Discussion on Style-Transfer Methods.} Regarding the alternative of style-transfer methods: these approaches typically require an additional style reference image and a dedicated style-transfer model (either CNN-based or diffusion-based) to generate stylized outputs, which introduces comparable or even greater complexity. More critically, most style-transfer methods alter not only the background but also distort the appearance of foreground objects, making them unsuitable for learning background-invariant representations. Some methods also require re-segmentation (e.g., using SAM) on the stylized images to recover object masks, which introduces noise and inconsistency—as discussed in Sect.\ref{IMBC}. This is especially unacceptable for object detection tasks, where precise annotations are essential.

Most importantly, typical style-transfer methods take both a content image and a style image as input, and may partially remove foreground objects during the transfer process. While some mask-aware methods attempt to mitigate this by treating the task as image inpainting, we have analyzed the theoretical limitations of applying inpainting-based techniques within our pipeline in the main paper, and provided supporting experimental results in Sect.\ref{IMBC}. These results show that inpainting methods perform poorly in our context.

\begin{table}[ht]
\centering
\caption{Comparison Between Our Method and Simple BBox Copy-Paste Based on the FIBER Baseline}
\resizebox{\linewidth}{!}{ 
\begin{tabular}{c c c c c c c}
\toprule
Method & AP & Ap-c & Ap-d & FULL & PRES & ABS \\
\midrule
FIBER-B+ours              & 42.0 & 44.1 & 39.2 & 37.6 & 37.2 & 38.8 \\
FIBER-B + BBox copy-paste & 23.3 & 27.0 & 20.6 & 20.9 & 20.4 & 22.6 \\
\bottomrule
\end{tabular}
}
\label{copypaste}
\end{table}

\subsection{Analysis on CBDG Scene Diversity}
\label{SceneDiversity}
While our current prompt set primarily focuses on Seasonal, Sky, and Natural Landscape categories, our goal is to begin with visually clean and semantically neutral backgrounds in order to isolate the effect of background variation on object-level consistency.

We conduct an experiment targeting urban, indoor, and architectural scenes. Specifically, we extract a subset of images from the $D^3$ test set that depict cityscapes and indoor urban environments. This subset includes the following categories: accidents, apples, auto salon girls, bakery, bananas, bedroom, boxing, cable car, cafe, chess, classroom, cooking, dining table, gymnastics, interview, kendo, library, living room, meeting, Olympic torch, origami, oven kitchen, pizza restaurant, refrigerator, restaurant, sandwich restaurant, snooker, street performers, street vendors food carts, study, supermarket cart, surgery hospital, toilet toothbrush, toy, traffic, and traffic accident, totaling 2,304 images.

We evaluate our model on this subset. As shown in Table~\ref{urban}, our method achieves results comparable to those on the full $D^3$ test set. This indicates that the learned representations generalize well even to previously less frequent scene types.

\begin{table}[ht]
\centering
\caption{Results on 2,304 $D^3$ Images from Cityscapes and Indoor Urban Environments}
\begin{tabular}{c c c c}
\toprule
Method & FULL & PRES & ABS \\
\midrule
GLIP-T+ours  & 29.7 & 28.9 & 32.1 \\
FIBER-B+ours & 37.5 & 37.1 & 38.6 \\

\bottomrule
\end{tabular}
\label{urban}
\end{table}

\subsection{Experiments on phrase grounding}
\label{discussiona}
To test whether our fine-tuning affects the baselines’ object grounding capabilities, we revisit a key benchmark used in both the GLIP and FIBER papers: phrase grounding on the Flickr30K Entities dataset. This benchmark requires precise spatial localization of objects given natural language descriptions, and thus reflects the model's fine-grained visual grounding ability. We report Recall@K metrics for both GLIP-T and FIBER-B before and after applying our method:

\begin{table}[ht]
\centering
\caption{Recall@K metrics for both GLIP-T and FIBER-B before and after applying our method}
\resizebox{\linewidth}{!}{ 
\begin{tabular}{c ccc ccc}
\toprule
\multirow{2}{*}{Model} & \multicolumn{3}{c}{Val} & \multicolumn{3}{c}{Test} \\
\cmidrule(lr){2-4} \cmidrule(lr){5-7}
 & R@1 & R@5 & R@10 & R@1 & R@5 & R@10 \\
\midrule
GLIP-T       & 85.7 & 95.4 & 96.9 & 85.7 & 95.8 & 97.2 \\
GLIP-T+ours  & 88.5 & 95.9 & 97.2 & 88.8 & 96.2 & 97.2 \\
FIBER-B      & 87.1 & 96.1 & 97.4 & 87.4 & 96.4 & 97.6 \\
FIBER-B+ours & 89.9 & 96.6 & 97.7 & 90.0 & 96.8 & 97.7 \\
\bottomrule
\end{tabular}
}
\end{table}

The performance is overall better than the original baselines. Notably, GLIP-T + ours achieves consistent gains on R@1 for both validation and test sets, while FIBER-B + ours also improves across several metrics. These results confirm that our consistency-driven fine-tuning not only preserves but can even enhance the grounding capabilities of the baselines.

It is notable that this evaluation serves as a preliminary validation, since our model is currently fine-tuned on only a 250K subset of the training data, which is significantly smaller than the full datasets used to train GLIP and FIBER. We expect that with larger-scale training, our approach has strong potential to further surpass the baselines on phrase grounding benchmarks like Flickr30K Entities. In general, these additional experiments demonstrate that our method maintains or improves the original capabilities of the baselines while simultaneously enhancing robustness under domain shifts.

\subsection{Analysis on Inpainting Methods for Background Changing}
\label{IMBC}
In our background replacement approach, we initially adopt different methods. We explore models such as Glide~\cite{nichol2021glide} and Inpaint Anything~\cite{yu2023inpaint} to integrate the original image with newly generated backgrounds using image-to-image techniques. Our process involves providing three inputs to the image-to-image model: the original image, the bounding box of the foreground object as a prompt, and a simple textual description of the desired background. The model then selectively modifies the non-foreground regions of the image.

However, the results are unsatisfactory, presenting three major issues, as shown in Figure~\ref{fig:inpaintfail}.a) Blurry Transition: the transition between the foreground and the generated background is often blurry, making the composite image appear unnatural. b) Extraneous Objects: the model tends to introduce extraneous objects into the background based on the foreground elements, which significantly disrupts the accuracy of subsequent object detection tasks. c) Logical Inconsistency: the generated images frequently contain logical inconsistencies that deviate from real-world plausibility, further diminishing their usability in our experiments.

We begin our investigation into synthetic data augmentation by utilizing the text-conditional inpainting capabilities of the GLIDE~\cite{nichol2021glide} model to modify image backgrounds according to textual prompts. However, initial experiments reveal that training with GLIDE-generated data leads to a notable drop in detection performance compared to the baseline, indicating limited utility of the inpainted samples. To improve data quality, we then experiment with the IAM~\cite{yu2023inpaint} model and introduce filtering strategies aimed at reducing noise and artifacts. It is worth noting that, unlike the experiments in the main paper, where we fine-tune with 0.25M images, here we used 0.09M images during this experiment to enable faster iteration at the early design stage. Consequently, the absolute performance numbers in Table~\ref{tabdgm} are lower than those in the main results, but the relative improvements between different generation methods remain valid. As presented in Table~\ref{tabdgm}, even with filtering, the addition of IAM-generated data still results in a performance decline. Nevertheless, across all settings, our proposed CCLoss consistently demonstrates its effectiveness in enhancing model robustness.

Inpainting-based approaches such as GLIDE and IAM frequently produce visual artifacts or semantically inconsistent elements in the generated backgrounds. These imperfections compromise data quality and limit generalization during training. To overcome these issues, we transition to using the Stable Diffusion model~\cite{rombach2022high} and adopt the full Context-aware Background Diversification and Generation (CBDG) pipeline detailed in our methodology. This revision yields a significant performance boost, affirming that high-quality synthetic data generated through our CBDG framework is critical for effective training and robust model behavior.




\begin{table}[ht]
\centering
\caption{Experiments on different generation method.}
\resizebox{\linewidth}{!}{ 
\begin{tabular}{c c c c |c c c}
    \toprule
    \multirow{2}{*}{Method(Baseline)} &\multicolumn{3}{c}{OmniLabel} &\multicolumn{3}{c}{$D^3$}\\
        &AP & AP-c & AP-d &FULL & PRES & ABS\\
    \midrule
    {IAM(FIBER-B)} &20.5&24.6&18.0     &18.4&17.8&20.3 \\
    {+ours(FIBER-B)} &26.1&31.2&22.2     &23.5&22.8&26.1 \\
    {CBDG+ours(FIBER-B)} & 32.4 & 34.6 & 29.4 & 30.3 & 30.0 & 31.4\\
    \bottomrule
\end{tabular}
}
\label{tabdgm}
\end{table}

\subsection{Progressive Design of the Intra-Modal Consistency Loss}
\label{PDICL}
The design of our intra-modal consistency loss is motivated by related work in self-supervised learning and domain adaptation, where preserving feature consistency under varying conditions has proven effective. In our framework, the development of the final loss function, CCLoss, follows a progressive and iterative design process. We begin by constructing a pairwise similarity matrix for the feature set $F={f_1, f_2, ..., f_n}$ corresponds to the same object instance presented under different background conditions. In the initial formulation, we minimize the mean squared error (MSE) between the lower triangular part of this similarity matrix and an all-ones target matrix, encouraging uniform similarity among all feature representations. This variant is referred to as {Matrix(FIBER-B)} in Table~\ref{tabdcl}. For each object in the same batch $n$, the loss is defined as follows: 

\begin{equation}
\mathcal{L}_{\text{Matrix}} = \frac{2}{n(n+1)} \sum_{i=1}^{n} \sum_{j=1}^{i} \left( \mathbf{S}(f_i, f_j) - 1 \right)^2,
\end{equation}

where $\mathbf{S}(f_i, f_j)$ represents the similarity between the feature vectors $f_i$ and $f_j$, which is computed using dot product.

To further enhance consistency, we incorporate the concept of variance reduction inspired by Variance Consistency Loss. Specifically, we minimize the variance among the features in $F$, promoting tighter feature clustering under intra-object background shifts. This second variant is denoted as {Variance(FIBER-B)} in Table~\ref{tabdcl}. For each object in the same batch $n$, the loss is defined as follows: 

\begin{equation}
\mathcal{L}_{\text{Variance}} = \frac{1}{n} \sum_{i=1}^{n} \left\| f_i - f_c \right\|^2,
\end{equation}

where $f_c$ is the centroid of the image features for the c-th category object.

Building upon these two approaches, we finally adopt a contrastive learning framework to enforce discriminative yet consistent representation learning. This final version, termed CCLoss, outperforms the earlier variants and demonstrates superior robustness across all experimental benchmarks. It is worth noting that, unlike the experiments in the main paper, where we fine-tune with 0.25M images, here we used 0.09M images during this ablation to enable faster iteration at the early design stage. Consequently, the absolute performance numbers in Table~\ref{tabdcl} are lower than those in the main results, but the relative improvements between different loss functions remain valid. The progressive improvements reflected in Table~\ref{tabdcl} validate the effectiveness of each stage in the design of our loss function.

\begin{table}[ht]
\centering
\caption{Experiments on different consistency loss.}
\resizebox{\linewidth}{!}{ 
\begin{tabular}{c c c c |c c c}
    \toprule
    \multirow{2}{*}{Method(Baseline)} &\multicolumn{3}{c}{OmniLabel} &\multicolumn{3}{c}{$D^3$}\\
        &AP & AP-c & AP-d &FULL & PRES & ABS\\
    \midrule
    {Matrix(FIBER-B)} &29.8&32.4&27.5     &27.9&26.8&30.0 \\
    {Variance(FIBER-B)} &30.2&33.2&28.6     &28.8&28.1&30.5 \\
    {CCLoss(FIBER-B)} & 32.4 & 34.6 & 29.4 & 30.3 & 30.0 & 31.4\\
    \bottomrule
\end{tabular}
}
\label{tabdcl}
\end{table}

\subsection{Analysis on Data Scale}
\label{datascale}
To further explore the impact of data scale, we conduct an additional experiment where we systematically modify the size of $D_j$. In this experiment, we reduce the dataset size and evaluate the model’s performance to determine how much the consistency-based approach can still contribute under these conditions. The reduction scale factor is $K_r$. As shown in Table~\ref{ds}, even when the dataset size is reduced by 0.8 or even 0.6, the consistency method continues to provide a substantial performance improvement, demonstrating that the proposed consistency approach remains effective even with limited data. This result underscores the robustness of our method. The findings suggest that our approach holds significant promise for scenarios where data availability is constrained, offering an effective solution to improve model generalization and performance even when working with smaller datasets. This insight could be particularly valuable in real-world applications where large-scale labeled data may not be easily accessible.

\begin{table}[t]
\centering
\caption{Impact of dataset size of the CCL approach, where $K_r$ is the dataset reduction scaling factor.}
\resizebox{\linewidth}{!}{ 
\begin{tabular}{c c c c |c c c}
    \toprule
    \multirow{2}{*}{Method(Baseline)} &\multicolumn{3}{c}{OmniLabel} &\multicolumn{3}{c}{$D^3$}\\
        &AP & AP-c & AP-d &FULL & PRES & ABS\\
    \midrule
    GLIP-T & 19.3 & 23.6 & 16.4 & 19.1 & 18.3 & 21.5\\
    +ours($K_r$=0.6) & 27.2 & 30.4 & 24.7 & 26.2 & 25.2 & 29.3\\
    +ours($K_r$=0.8) & 30.1 & 34.1 & 27.4 & 28.9 & 28.1 & 31.3\\
    \midrule
    FIBER-B & 25.7 & 30.3 & 22.3 & 22.7 & 21.5 & 26.0\\
    +ours($K_r$=0.6) & 35.6 & 38.3 & 34.0 & 32.4 & 32.0 & 33.6\\
    +ours($K_r$=0.8) & 39.5 & 41.6 & 36.7 & 35.8 & 35.5 & 37.0\\
    \bottomrule
\end{tabular}
}
\label{ds}
\end{table}

\section{Related Work}

\subsection{Robust Object Detection}
\label{ROD}
Robust object detection focuses on maintaining reliable performance under diverse challenging conditions, including occlusion, adverse weather, low-resolution inputs, domain shifts, and adversarial attacks. While traditional object detectors operate under idealized data assumptions, robust detection methods target the inherent variability of real-world environments. The field has seen significant progress through innovations in model architectures, training paradigms, and domain adaptation methodologies. Current research advances tackle robustness challenges through multiple directions: Adversarial training, where FROD~\cite{awais2023frod} improves detector robustness via modified backbones and lightweight components. Extreme condition adaptation, with UIA-YOLOv5~\cite{ding2024robust} enhancing construction site detection through unified image adaptation. Noisy bounding box handling, as OA-MIL~\cite{liu2022robust} refines localization using classification guidance. Domain generalization, where Normalization Perturbation~\cite{fan2023towards} synthesizes feature styles for autonomous driving. Small object detection, with DenseNet-201~\cite{akhtar2022robust} boosting YOLOv2 for traffic surveillance. Agricultural robustness, where smooth perturbations~\cite{mahmoud2024robust} improve YOLOv5 for root collar detection. These methods span adversarial resilience, environmental adaptability, and annotation noise tolerance.

\subsection{Data Augmentation for Detectors}
\label{DAD}
Recent advancements in detector-specific data augmentation address limitations in classic methods like Mixup~\cite{zhang2017mixup} and CutMix~\cite{yun2019cutmix}, which primarily enhance diversity in fully supervised settings. Diffusion-based frameworks generate diverse contrail masks and scenes to improve detection robustness~\cite{lee2025improving}, while attribution-driven methods leverage saliency maps to preserve critical features in low-level vision tasks~\cite{mi2025add}. For few-shot object detection, MPAD~\cite{vu2025multi} integrates in-context object synthesis and hard sample generation via diffusion models. Nevertheless, persistent challenges—such as edge fidelity issues (e.g., blurred boundaries in diffusion-generated regions) and computational inefficiency—have led to renewed interest in hybrid augmentation strategies combining classical and modern techniques. Our proposed CBDG method uniquely integrates SAM (Segment Anything Model) and Stable Diffusion to generate semantically coherent, background-diverse images while minimizing label noise for consistency training. Unlike prior augmentation approaches, CBDG is designed to facilitate robust object-level feature learning across diverse environments in open-vocabulary scenarios, effectively mitigating common artifacts such as blurred boundaries.

\subsection{Cross-Modal Object Detection Models}
\label{CMODM} 
Cross-modal object detection develops rapidly with advances in vision–language pretraining and grounding. Early frameworks such as CLIP~\cite{radford2021learning} and ALIGN~\cite{jia2021scaling} establish the foundation by learning scalable image–text representations, which inspire a series of detection-oriented extensions. MDETR~\cite{kamath2021mdetr} aligns objects and textual queries through a transformer-based design, while OWL-ViT~\cite{minderer2022simple} enables open-vocabulary detection with vision transformers. Subsequent models, including G-DINO~\cite{liu2024grounding} and OFA-DOD~\cite{xie2023described}, further advance grounding accuracy and generalization across diverse tasks. Among these, GLIP~\cite{li2022grounded} and FIBER~\cite{dou2022coarse} emerge as representative and widely adopted frameworks. GLIP introduces a scalable region-level vision–language pretraining formulation, whereas FIBER adopts a two-stage coarse-to-fine strategy that improves alignment at both image and region levels. Both models provide strong baselines for language-supervised object detection, support text-based grounding tasks, and are publicly available, which makes them ideal foundations for evaluating and extending consistency-based improvements. Their popularity and architectural compatibility also highlight the model-agnostic nature of our proposed approach.

\subsection{Survey of Existing Background Datasets}
\label{existingdata}

To assess the suitability of existing datasets for our method, we conduct a detailed survey of publicly available background datasets. We find that nearly all of them contain varying degrees of foreground objects, which fundamentally conflicts with our training objective: pasting finely segmented foregrounds (via SAM) onto clean backgrounds to enforce background-invariance.

\paragraph{Foreground contamination.} Many background datasets include objects that should not appear in background-only scenes. For instance, when sampling 200 random images from BG-20k~\cite{li2022bridging}---often cited as a background-only dataset---we found that 36\% contained recognizable foregrounds such as roses, dolphins, butterflies, and other clearly foreground elements. Other datasets (e.g., SUN09~\cite{xiao2010sun}, Stanford Background~\cite{gould2009decomposing}, Cityscapes~\cite{cordts2016cityscapes}) often include pedestrians, vehicles, or other labeled objects, introducing semantic confusion. This undermines the consistency objective by entangling background pixels with foreground semantics.

\paragraph{Domain-specificity.} Some datasets (e.g., NH-HAZE~\cite{ancuti2020nh}, RESIDE~\cite{li2018benchmarking}) are restricted to specific conditions such as hazy weather, making them unsuitable for general-purpose training.

\paragraph{Scale limitations.} Many datasets are too small to support large-scale pretraining or fine-tuning. For example, the Stanford Background Dataset contains only 715 images, Cityscapes provides 2,975 training images within a single urban domain, and NWPU VHR-10 has merely 150 background images.

For methods that explicitly separate foreground and background, dataset purity is crucial. Existing background datasets either suffer from foreground contamination, domain bias, or insufficient scale, and therefore cannot fully meet the requirements of our approach.

\section{CBDG}

\subsection{Semantic Conflicts and Object-Background Decoupling}
\label{semanticD}
\begin{figure*}[t]  
    \centering
    \includegraphics[width=\textwidth]{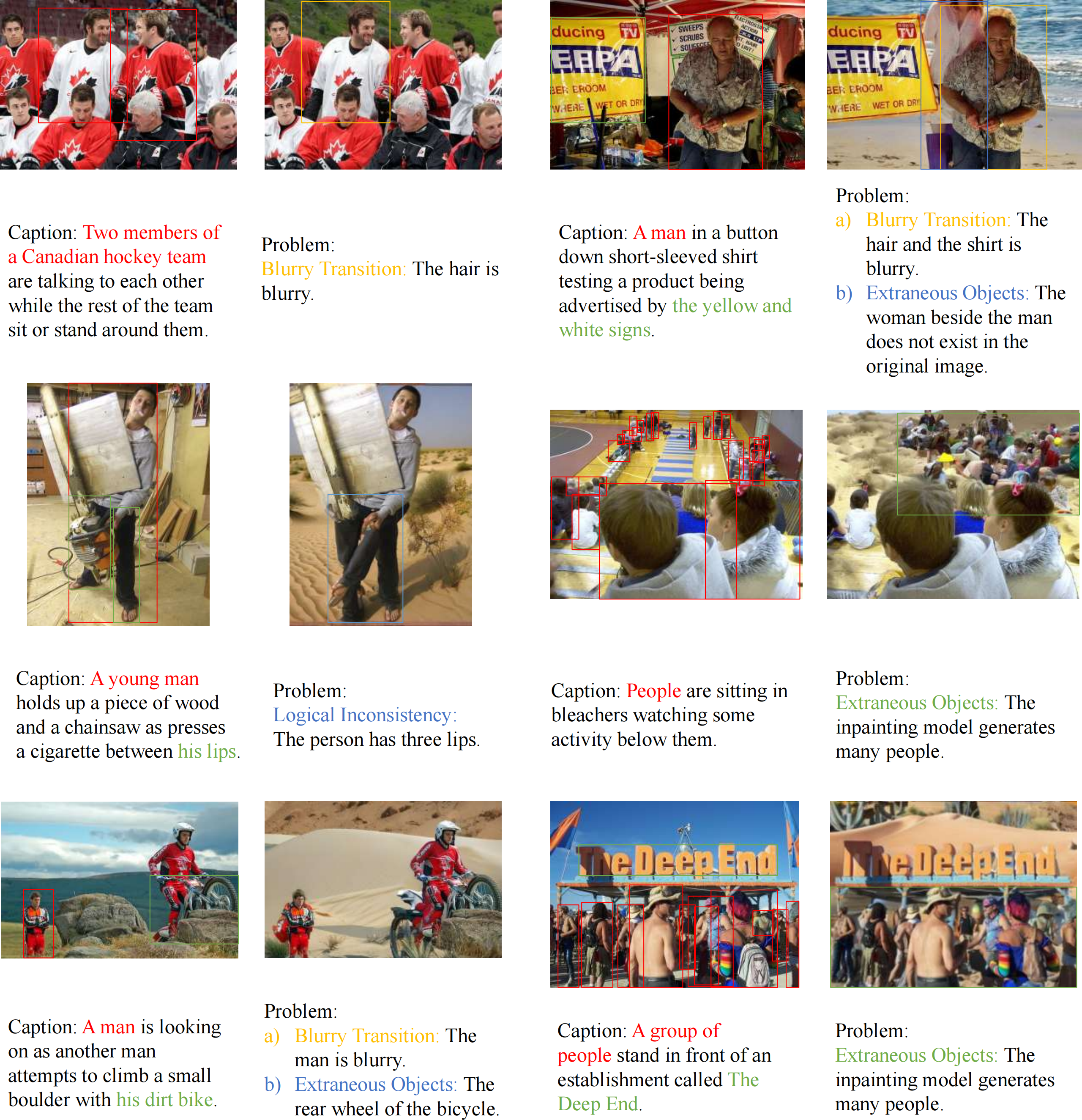}  
    \caption{Failure cases of background replacement using inpainting method~\cite{yu2023inpaint}. Each case consists of paired images: the left image shows the original input, while the right image demonstrates the replaced background with corresponding artifacts. Cases are presented in a two-per-row layout for comparison.}
    \label{fig:inpaintfail}
\end{figure*}

While natural object–background correlations exist, they often induce over-reliance on spurious contextual cues. \textbf{CBDG intentionally decouples objects from their original backgrounds to promote object-centric and robust representations.} Although this may introduce semantically uncommon combinations, our model consistently outperforms baselines on $D^3$, $D^3_{BC}$ and $D^3_{\mathrm{C}}$, indicating improved generalization rather than degraded learning. Potential failure cases from imperfect extraction are mitigated via post-processing filtering (Supplementary Section~\ref{EITTQ}) and visualized in Figure~\ref{fig:inpaintfail}.

\subsection{Prompt to LLM for Background Generation}
\label{promptLLM}

\begin{lstlisting}[
    caption={Example prompts for generating natural landscape background images with seasonal variations, time of day, and weather conditions.},
    label={code:prompt_example},
    breaklines=true,  % 启用自动换行
    postbreak=\mbox{\textcolor{red}{$\hookrightarrow$}\space}, % 可选：换行标记
    basicstyle=\small\ttfamily
]
Generate a list of diverse prompts for background image generation, focusing on natural landscapes (sky and atmosphere) with seasonal variations. Provide varied descriptions covering different times of day, weather conditions, and visual elements. The prompts should be clear, creative, and specific for use in diffusion models.

Generated examples:
1. A vibrant spring meadow with tall grass and colorful wildflowers, where bees and butterflies flit from bloom to bloom under the warm sun.
2. A frozen river cutting through a snowy valley, with ice floes floating on the surface and snow-capped trees lining the banks.
3. A tropical waterfall plunging down into a crystal-clear pool, surrounded by dense jungle and mist rising from the water below.
4. A golden autumn forest at sunset, with leaves in shades of red and orange covering the ground and a crisp chill in the air.
5. A stormy summer beach scene with dark clouds overhead, waves crashing against the shore, and seagulls fighting the strong winds.
6. A misty morning in a bamboo forest, with soft sunlight filtering through the tall green stalks and dew on the leaves.
7. A panoramic view of rolling hills under a twilight sky, with the first stars appearing and a crescent moon rising.
8. A desert oasis at high noon, with palm trees surrounding a small pool of water and heat waves distorting the distant horizon.
9. A cherry blossom grove in full bloom, with pink petals floating on a gentle breeze and covering a quiet pathway.
10. A northern lights display over a frozen lake, with vibrant green and purple streaks reflected in the icy surface.
\end{lstlisting}

We employ three distinct prompts to generate background descriptions for three different categories using ChatGPT. While the templates for these prompts are identical, they vary slightly in terms of the generated content, as illustrated in Listing~\ref{code:prompt_example}. The primary difference lies in the type of content generated, tailored to each category.

\subsection{Effect of IoU Thresholding on Training Quality}
\label{EITTQ}

To address the issue of incomplete or inaccurate foreground object segmentation by SAM, which is unacceptable for downstream object detection tasks, we implement a post-processing step to filter out problematic segmentation results. Specifically, we calculate the bounding box (BBox) of each segmented mask and compare it with the corresponding ground truth BBox by computing the Intersection over Union (IoU). Only segmentation results with an IoU greater than a predefined threshold $T_{IoU}$ are retained. Otherwise, they are discarded. The threshold $T_{IoU}$ we take is 0.75.

This post-processing step mitigates the impact of noisy BBox labels caused by potentially inaccurate SAM-generated masks, thereby improving the quality of training data and enhancing model performance. 



\subsection{Analysis on the Processing Procedures of Different Datasets} 
\label{PPDD}
In our CBDG process, we apply two different augmentation strategies to our combined dataset of Flickr30k and Objects365. For images with more than one class in them, we leverage the prior knowledge of category labels and bounding box information to directly extract the foreground objects using the SAM model. Once the foreground is segmented, we proceed to replace the background, ensuring that the core object remains intact while the surrounding context is altered. This method allows us to generate diverse augmented images while maintaining object consistency.

For images with only one class in them, we begin by augmenting the number of object categories within the images, expanding the variety and complexity of the scenes. After this, we apply the same background replacement technique as used for the Flickr30k dataset, using the extracted foreground objects and swapping the background accordingly. This dual augmentation approach helps us enrich the dataset, enhancing its diversity and challenging the model to adapt to a wider range of visual contexts. By utilizing both the category label and bbox priors in conjunction with the SAM model, we can effectively augment the datasets in a way that introduces meaningful variation while preserving object relevance, which is crucial for improving model robustness and generalization across different scenarios.

In the process of category augmentation, to minimize the impact of dataset size, the augmented objects are exclusively selected from 0.22M images from the Objects365 dataset, excluding objects from the images we did not select within the original Objects365 dataset.

\begin{figure}[t]  
    \centering
    \includegraphics[width=\columnwidth]{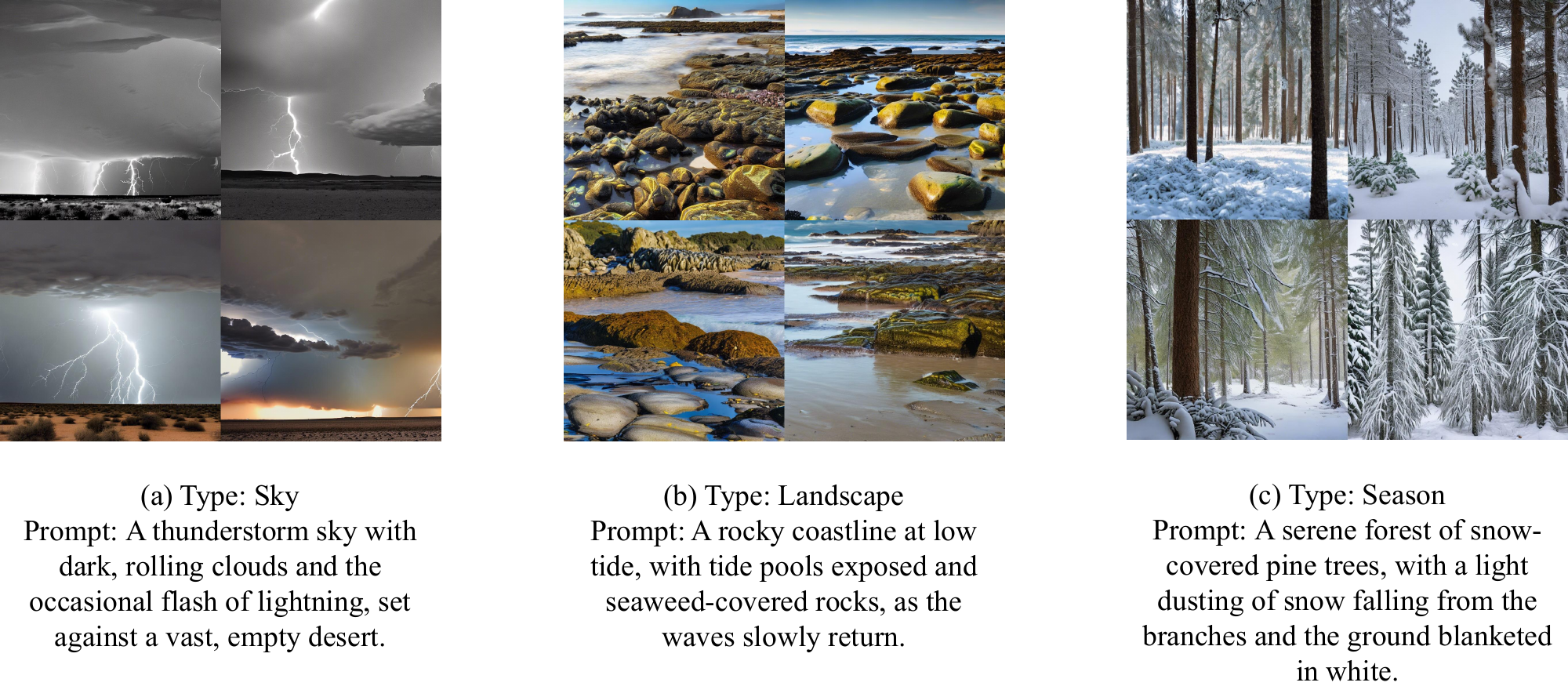}  
    \caption{Three types of background images we generate. Given a prompt from LLM, we use stable diffusion to generate random background images.}  
    \label{fig:genim}  
\end{figure}

\subsection{Analysis on Background Generation}
\label{ABG}
As shown in Figure~\ref{fig:genim}, the background images we generate are both rich and diverse, with significant variation in elements such as scene types, lighting conditions, weather changes and texture details. These images are created by providing prompts derived from various language descriptions generated by a large language model (LLM) to the Stable Diffusion model. The prompts are designed to capture different scenes, ensuring that the generated backgrounds exhibit a high degree of diversity.

There are three main categories of background images: seasonal, sky, and natural landscape. In total, we have 13,185 unique descriptions, resulting in 144,654 generated images. The breakdown of categories and the corresponding number of images is as follows: seasonal (3387 descriptions, 48,156 images), sky (3399 descriptions, 48,210 images), and natural landscape (3399 descriptions, 48,288 images). As shown in Figure~\ref{fig:genim}, even for the same language description, the generated images demonstrate a strong diversity in terms of their visual appearance, making each one unique in its own right.

This diversity is key to enhancing the quality and variability of the generated datasets, allowing us to better capture the complexity of real-world scenes and improve the robustness of subsequent tasks, such as OVOD and REC. Moreover, the diversity in the backgrounds plays a crucial role in supporting the consistency constraints we impose on the images. Unlike traditional data augmentation methods, which may lead to homogenized backgrounds, our approach ensures that the model is exposed to a wide range of backgrounds across different contexts. This enhances the model's robustness, making it more adaptable and effective in diverse scenarios, and ultimately improving its performance on tasks that require consistency across varying environments.

\subsection{Analysis on Image Selection and Categorical Augmentation}
\label{ISandCA}
For the Objects365 dataset, we initially select 0.22M images, ensuring that all categories are covered to support more comprehensive model training and evaluation. During the data selection process, we track the categories and the corresponding number of objects (bounding boxes) in each image. To ensure category balance, we first analyze the distribution of categories across the dataset and set a minimum coverage requirement, ensuring that each category is represented by at least one image. 

Subsequently, we introduce a priority strategy based on category diversity and object count to sort and select the images. The core idea behind this priority strategy is to prioritize images that contain a larger number of categories while favoring those with a higher number of objects. This approach allows for more efficient category coverage within a limited dataset and ensures that the dataset has an adequate number of objects to support robust object detection model training.

For each image, we define its priority score as:
\begin{equation}
    S(I) = N_{\text{bbox}}(I) + N_{\text{cat}}(I)
    \label{eq:datachoose}
\end{equation}
where $S(I)$ represents the priority score of image $I$, $N_{\text{bbox}}(I)$ denotes the number of objects (bounding boxes) in image $I$, $N_{\text{cat}}(I)$ refers to the number of categories in image $I$.

We sort the images by priority score and gradually select them. While ensuring category coverage, we further optimize the data selection based on object count and category diversity, such that the final subset contains both rich category information and reasonable object density. This strategy significantly enhances the representativeness of the dataset, providing more challenging and generalizable training data for subsequent object detection tasks.




\subsection{Pseudo Code of CBDG}
\label{pseudo code}
The overall process of contextual bootstrapping data generation is summarized in the Algorithm~\ref{alg:datageneration}.

\begin{algorithm}
    \caption{Contextual Bootstrapping Data Generation}
    \label{alg:datageneration}
    \begin{algorithmic}[1]
        \REQUIRE 
        \STATE $I$: Original image with objects and bounding boxes.
        \STATE $D_{fg}$: External dataset for object augmentation.
        \STATE $N$: Number of potential placement positions.
        \STATE $N_R$: Maximum resizing attempts.
        \STATE $\alpha$: Resizing factor.
        \STATE $\text{Themes}$: Background themes (\textit{Seasonal, Sky, Natural Landscape}).
        
        \ENSURE 
        \STATE $I^*$: Augmented image with diverse objects and backgrounds.
        
        \STATE Extract object using SAM $\mathcal{S}$: 
            $\text{foreground objects} \gets \mathcal{S}(I)$.
        \FOR{each object in $I$}
            \STATE Randomly select an object $o$ from $D_{fg}$.
            \STATE Attempt placement at $P$ positions:
            \STATE \quad $(x_{o_k}, y_{o_k})\in P\backslash(x_o,y_o), o_k \in O_{i\notin C}$.
            \STATE category augmentation: $I'\gets I$
            \IF{no valid placement found}
                \STATE Resize $o$: $o \gets o \times 1/\alpha$.
                \STATE Repeat placement attempts up to $N_R$ times.
                \IF{still no valid placement}
                    \STATE Skip and select another image.
                \ENDIF
            \ENDIF
        \ENDFOR
        \STATE Generate background prompts using LLM $\mathcal{G}$:
        \STATE \quad $t' = \mathcal{G}(t)$.
        \STATE Generate background images using Stable Diffusion $\mathcal{D}$:
        \STATE \quad $b=\mathcal{D}(t')$.
        \STATE Replace background in $I'$:
        \STATE \quad Extract foreground and compose new image: $I^*=\mathcal{S}(I'_{box})\oplus b, b\in D_{bg}$.
        \STATE \RETURN $I^{*}$.
    \end{algorithmic}
\end{algorithm}